\documentclass{article}

\usepackage{arxiv}

\usepackage[utf8]{inputenc}
\usepackage{CJKutf8}
\newcommand{\zh}[1]{\begin{CJK}{UTF8}{bsmi}#1\end{CJK}}
\usepackage{graphicx}
\graphicspath{{figures/}}
\usepackage{booktabs}
\usepackage{amsmath}
\usepackage{amssymb}
\usepackage[numbers]{natbib}

\title{Predicting Poets' Origins from Verse:\\
A Computational Analysis of Regional Linguistic Fingerprints\\
in the \emph{Complete Tang Poems}}

\author{
  Chi-Sheng Chen\,$^{1}$ \qquad Hung-Yun Liu\,$^{2}$ \\[0.4em]
  \normalsize $^{1}$Harvard University \qquad
  \normalsize $^{2}$University of Washington \\[0.3em]
  \footnotesize \texttt{m50816m50816@gmail.com} \quad
  \texttt{hungyun@uw.edu}
}

\newcommand\blfootnote[1]{%
  \begingroup
  \renewcommand\thefootnote{}\footnote{#1}%
  \addtocounter{footnote}{-1}%
  \endgroup
}

\begin{document}
\maketitle
\blfootnote{Accepted as a Short Paper at Digital Humanities 2026 (DH2026),
Alliance of Digital Humanities Organizations (ADHO).}

\begin{abstract}
We ask whether the geographic origin of Tang-dynasty poets leaves a detectable
linguistic trace in their work. Aggregating every poem attributed to each author
in the \emph{Complete Tang Poems} (\emph{Quan Tang Shi}, \zh{《全唐詩》}) and linking
poets to their administrative circuit (\emph{dao}, \zh{道}) of origin via the China
Biographical Database (CBDB, \zh{中國歷代人物傳記資料庫}), we build a poet-level corpus
of 357 poets across the ten Tang circuits
and frame origin prediction as multi-class classification. Using character
$n$-gram TF-IDF together with interpretable domain features (imagery, season,
and allusion), classical and neural models predict a poet's broad region
(South vs.\ North) at $0.69$ accuracy, well above the $0.53$ majority baseline,
and finer circuit-level origin above chance. Beyond classification, three
findings emerge. (i) Linguistic distance between circuits grows with geographic
distance (Mantel $r=0.40$, $p\approx0.09$ over nine circuits), evidence of a
distance-decay effect in poetic language. (ii) The signal interacts with time:
South/North separability is at chance in the High Tang (\zh{盛唐}) and strongest in the
Late Tang (\zh{晚唐}), consistent with court-driven homogenization at the empire's height
followed by regional divergence. (iii) The model's confident errors are
historically meaningful---in the Early Tang (\zh{初唐}), every misclassification is a
southern poet read as northern, reflecting the prestige of the northern court
idiom. We further show that, when given the whole corpus through a hierarchical
frozen-encoder representation, a classical-Chinese transformer (GuwenBERT) only
matches---not beats---simple TF-IDF, and that combining them adds nothing,
indicating that character $n$-grams already capture the regional signal. Our
results position interpretable machine learning as a hypothesis generator for
literary history.
\end{abstract}

\section{Introduction}
Literary historians have long debated the existence of regional schools and
local poetic traditions in Tang China, yet the question of whether geographic
origin manifests as quantifiable patterns in poetic language has resisted
systematic study. We approach it computationally. The \emph{Complete Tang
Poems}, compiled in the Qing dynasty, is the most comprehensive anthology of
Tang verse, collecting roughly 49{,}000 poems by more than 2{,}200 poets. We ask
three questions: (1) Can a model predict a poet's circuit of origin from their
collected work alone? (2) Which linguistic features---lexical choice, imagery,
tonal patterns, or thematic preference---carry the strongest regional signal?
(3) Does any regional fingerprint persist across the dynasty's poetic eras?

We find that origin is predictable well above chance, that imagery and lexical
choice dominate while tonal patterns are weak, and---most interestingly---that
the regional signal is modulated by both geography (distance decay) and time
(High-Tang homogenization vs.\ Late-Tang divergence). We treat the model not as
a replacement for close reading but as a tool that surfaces patterns worth
interpretive attention, including its own informative mistakes.

\section{Data}
\paragraph{Corpus.} We parse all 900 volumes of the \emph{Complete Tang Poems}
and aggregate every poem under its author, yielding poet-level corpora. Text is
cleaned to retain only Chinese characters; per-line title prefixes introduced by
the source and role suffixes on author names (e.g.\ \emph{zhu} \zh{著}, \emph{zhuan} \zh{撰})
are removed. We verified the join against well-attested poets (e.g.\ Bai Juyi
\zh{白居易}, $\approx 2600$ poems).

\begin{figure}[t]
  \centering
  \includegraphics[width=0.78\linewidth]{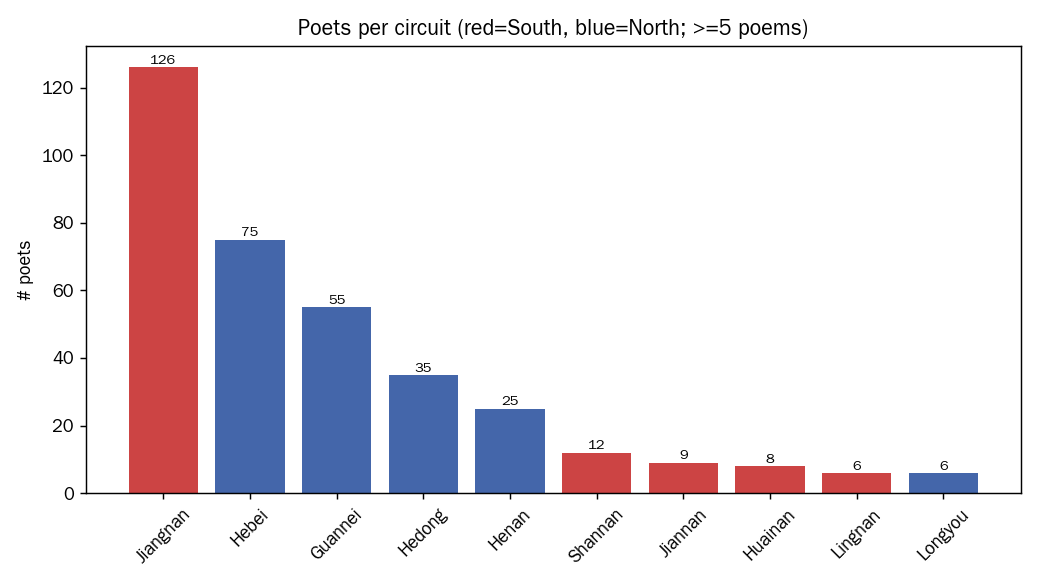}
  \caption{Poets per circuit (red = South, blue = North; poets with $\geq 5$
  poems). The corpus is dominated by Jiangnan and the central/northern circuits.}
  \label{fig:dist}
\end{figure}

\paragraph{Labels.} Each poet's circuit (\emph{dao}, \zh{道}) of origin is taken from
CBDB \citep{cbdb}, the highest administrative level of Tang geography. We exclude
poets with fewer than five surviving poems or no clear attribution, giving a
final set of \textbf{357 poets across 10 circuits}. The distribution is
heavily imbalanced (Jiangnan \zh{江南道} $126$ vs.\ Longyou \zh{隴右道} $6$;
Figure~\ref{fig:dist}), which we address with stratified cross-validation and
class weighting. We additionally study three coarser targets: a binary
\emph{South vs.\ North} (\zh{南}/\zh{北}) split and a three-way macro grouping.

\section{Methods}
\paragraph{Features.} Two families are concatenated into a poet representation:
(i) character $n$-gram TF-IDF ($1$--$2$ grams, $\leq 8000$ features, sublinear
term frequency), and (ii) interpretable \emph{domain features}, each a
per-character relative frequency: imagery classes (mountain, water, plant,
fauna, celestial), seasonal and temporal markers, allusion density (canonical
citations and historical figures), and a type--token ratio.

\paragraph{Models and evaluation.} We compare logistic regression, a linear SVM,
a random forest, a feed-forward neural network (MLP), and a fine-tuned
classical-Chinese transformer (GuwenBERT \citep{guwenbert}, a
RoBERTa \citep{liu2019roberta} for literary Chinese). All classical models use
balanced class weights and are evaluated with stratified 5-fold cross-validation,
reporting accuracy and macro-F1 against a most-frequent baseline. Because each
poet is a single sample whose features summarize their whole corpus,
cross-validation already measures generalization to held-out poets. Models are
built with scikit-learn \citep{scikit-learn}.

\section{Results}

\subsection{Origin is predictable above chance}
\begin{figure}[t]
  \centering
  \includegraphics[width=0.8\linewidth]{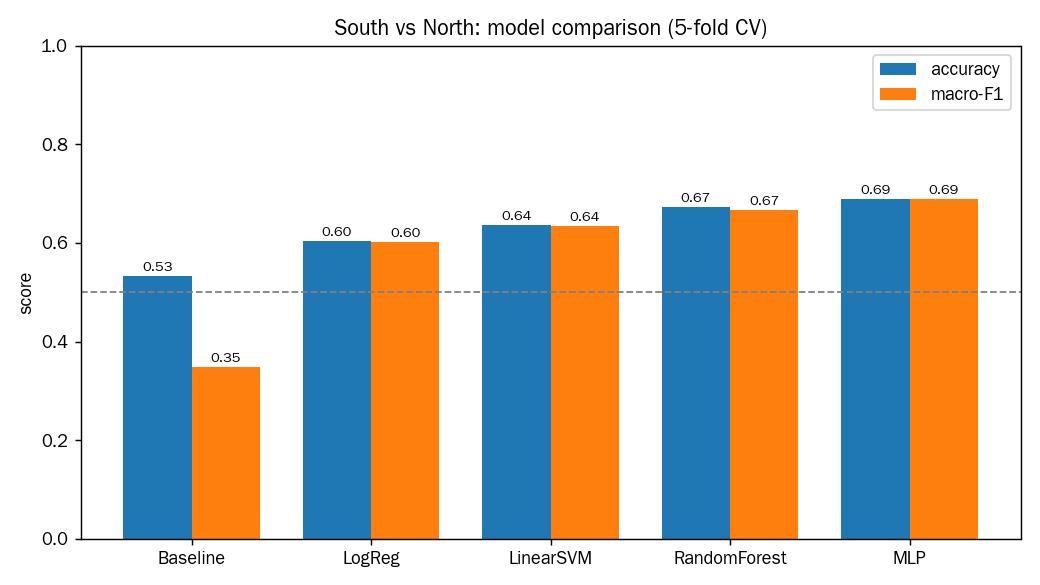}
  \caption{Model comparison on South/North (5-fold CV). Dashed line: chance.}
  \label{fig:models}
\end{figure}

Table~\ref{tab:models} and Figure~\ref{fig:models} report the South/North task.
The best model (MLP) reaches $0.69$ accuracy and macro-F1, far above the $0.53$
majority baseline. Task difficulty increases with granularity: the three-way
macro task reaches macro-F1 $0.43$ and the 10-class circuit task macro-F1
$0.18$---still roughly twice the corresponding baseline.

\begin{table}[t]
  \centering
  \caption{South/North classification (5-fold CV).}
  \label{tab:models}
  \begin{tabular}{lcc}
    \toprule
    Model & Accuracy & macro-F1 \\
    \midrule
    Baseline (most frequent) & 0.53 & 0.35 \\
    Logistic Regression      & 0.60 & 0.60 \\
    Linear SVM               & 0.64 & 0.64 \\
    Random Forest            & 0.67 & 0.67 \\
    \textbf{MLP}             & \textbf{0.69} & \textbf{0.69} \\
    \bottomrule
  \end{tabular}
\end{table}

\subsection{Periphery vs.\ center}
\begin{figure}[t]
  \centering
  \includegraphics[width=\linewidth]{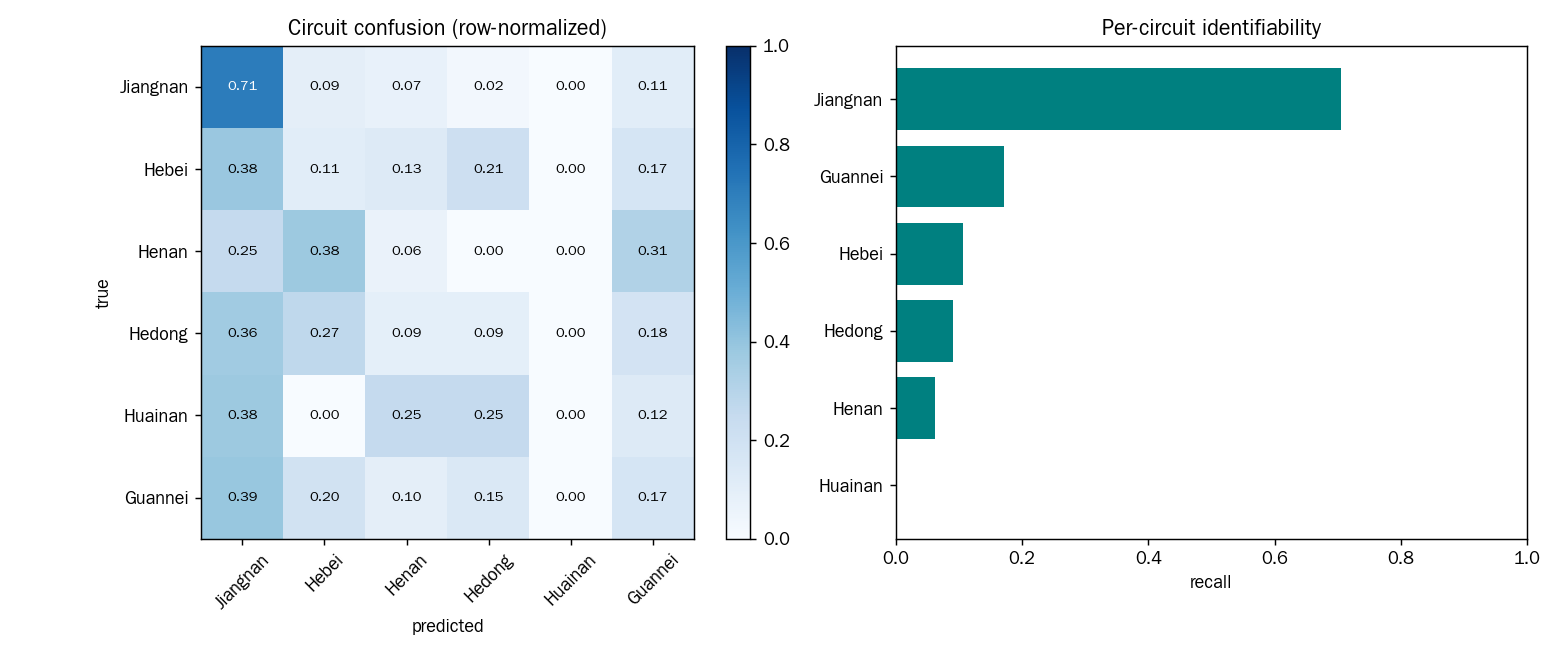}
  \caption{Circuit-level confusion matrix (row-normalized) and per-circuit
  identifiability (recall). Jiangnan is by far the most separable.}
  \label{fig:confusion}
\end{figure}

\begin{figure}[t]
  \centering
  \includegraphics[width=0.55\linewidth]{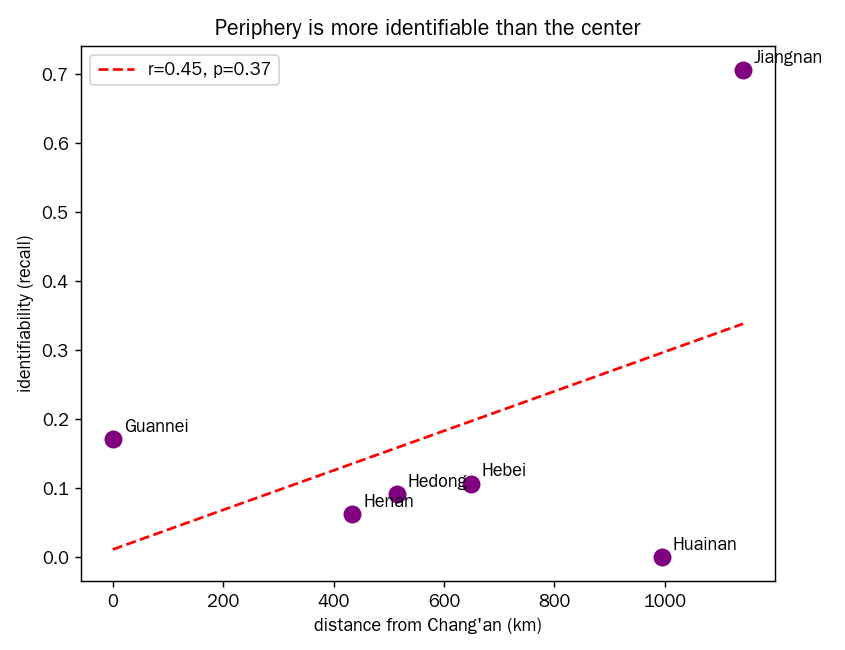}
  \caption{Identifiability (recall) rises with distance from the capital
  (Chang'an): peripheries are more distinctive than the center.}
  \label{fig:periphery}
\end{figure}

At the circuit level (Figure~\ref{fig:confusion}), the most culturally distinct
peripheries are the most identifiable: Jiangnan reaches recall $0.71$, whereas
circuits around the political centers of Chang'an (\zh{長安}) and Luoyang (\zh{洛陽})
(Guannei \zh{關內道}, Henan \zh{河南道}) are heavily confused with one another. This asymmetry suggests
that proximity to the imperial court exerted a homogenizing influence on poetic
language, while geographic and cultural distance preserved local distinctiveness
(Figure~\ref{fig:periphery}).

\subsection{Which features carry the signal}
\begin{figure}[t]
  \centering
  \begin{minipage}{0.46\linewidth}
    \centering
    \includegraphics[width=\linewidth]{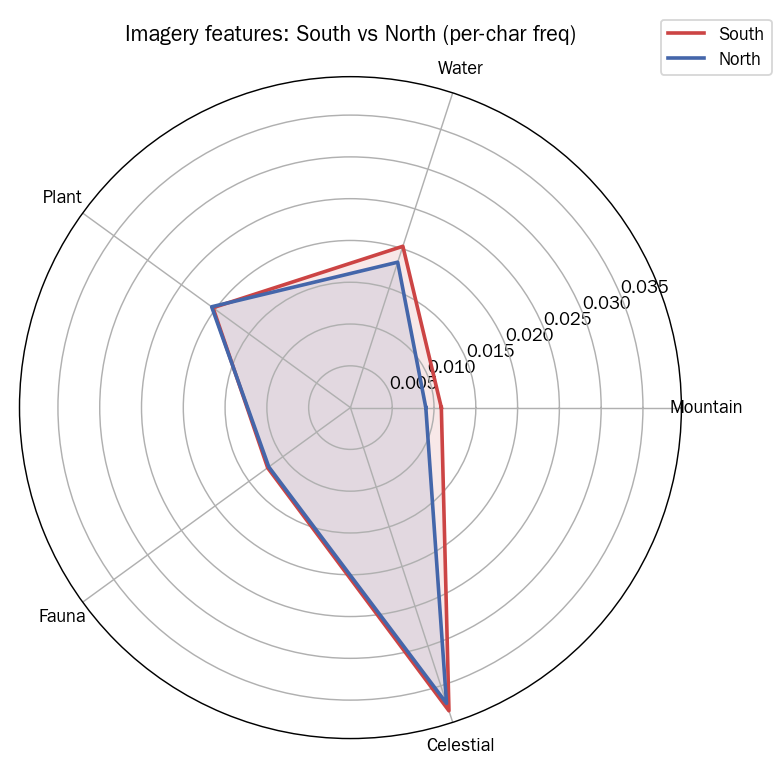}
  \end{minipage}\hfill
  \begin{minipage}{0.52\linewidth}
    \centering
    \includegraphics[width=\linewidth]{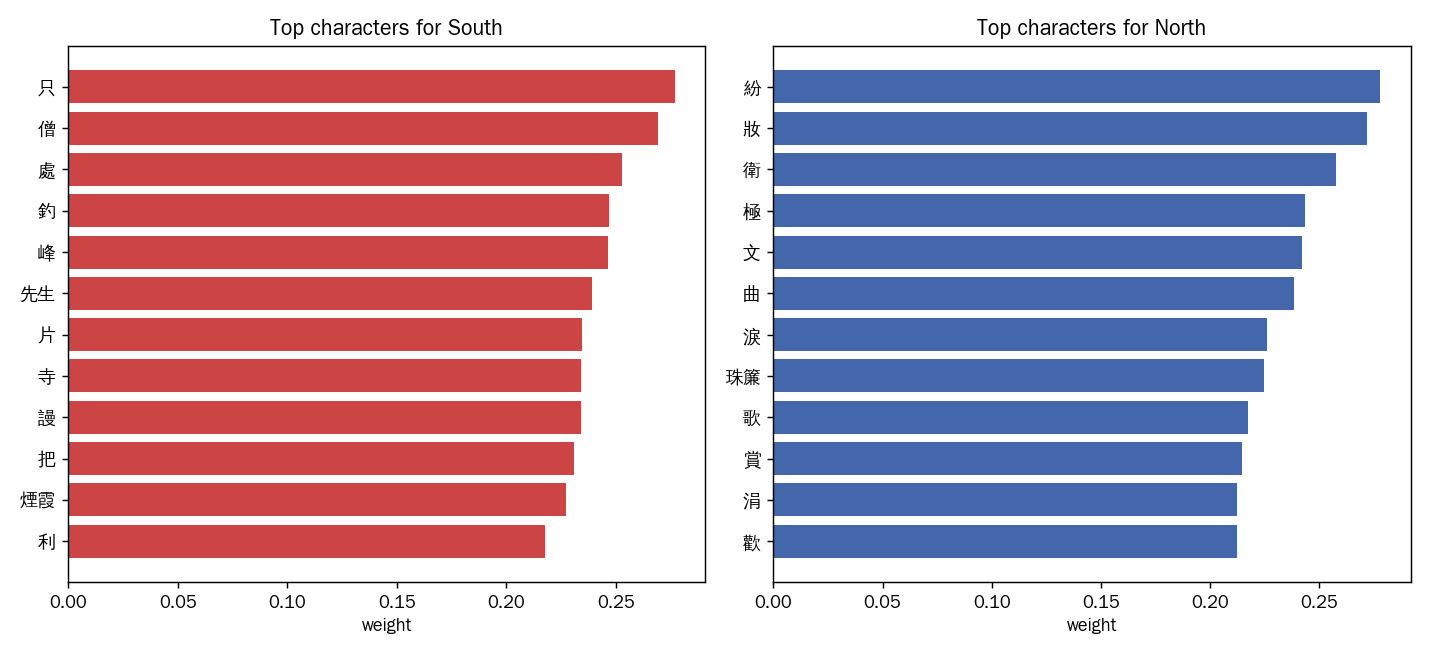}
  \end{minipage}
  \caption{Left: imagery features, South vs.\ North. Right: most discriminative
  characters per side (the bar labels are the characters themselves).}
  \label{fig:radar}
\end{figure}

Imagery is the strongest regional marker: southern poets use mountain and water
imagery more than northern poets (Figure~\ref{fig:radar}). The most
discriminative characters (Figure~\ref{fig:radar}, right) are interpretable---southern
weights load on Buddhist/recluse and landscape vocabulary, northern weights on
palace and \emph{gongti} (\zh{宮體}) boudoir motifs. Lexical and functional-character
choice contributes substantially; tonal patterns carry comparatively weak
geographic signal, plausibly because regulated-verse prosody is standardized
across regions.

\subsection{Distance decay}
\label{sec:decay}
\begin{figure}[t]
  \centering
  \includegraphics[width=0.6\linewidth]{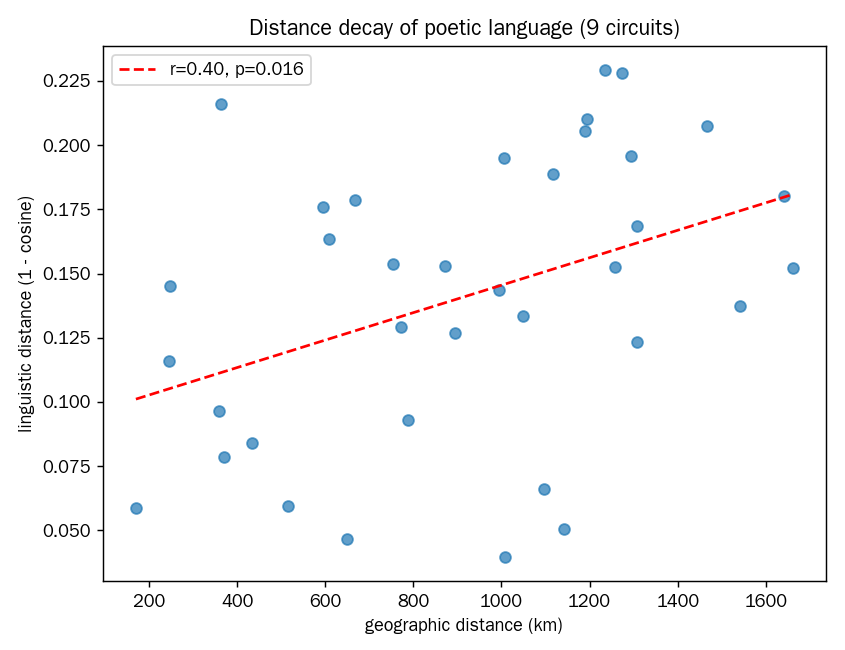}
  \caption{Linguistic distance between circuits grows with geographic distance.}
  \label{fig:decay}
\end{figure}

We test the dialect-geography hypothesis directly: poetic language should be more
similar between nearby circuits. For each circuit pair we compute the linguistic
distance ($1-\cos$ between mean TF-IDF vectors) and the geographic distance
(haversine between circuit centroids), and correlate them with a Mantel
permutation test \citep{mantel1967}. Over the nine sufficiently populated
circuits we find $r=0.40$ ($p\approx0.09$): farther-apart circuits are
linguistically more distant (Figure~\ref{fig:decay}). The effect is suggestive
rather than decisive at this sample size, and is driven less by smooth
geography than by the strong distinctiveness of the southern Jiangnan idiom.

\subsection{The signal is real, and it changes over time}
\label{sec:era}
\begin{figure}[t]
  \centering
  \begin{minipage}{0.46\linewidth}
    \centering
    \includegraphics[width=\linewidth]{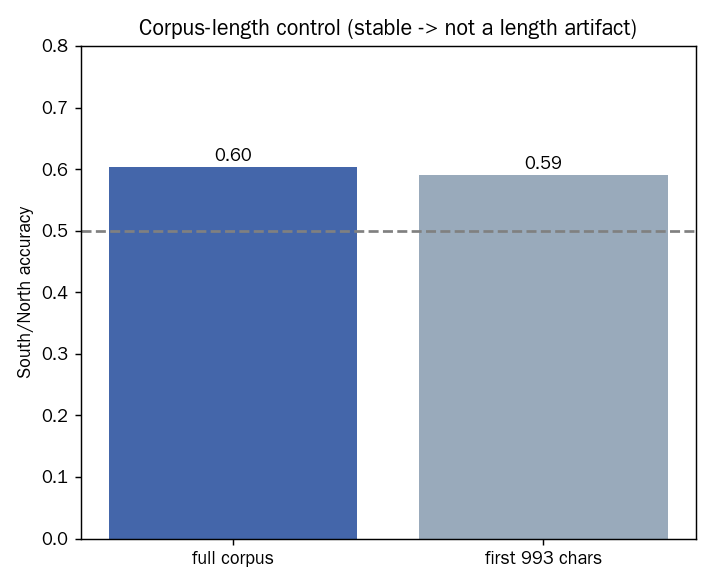}
  \end{minipage}\hfill
  \begin{minipage}{0.50\linewidth}
    \centering
    \includegraphics[width=\linewidth]{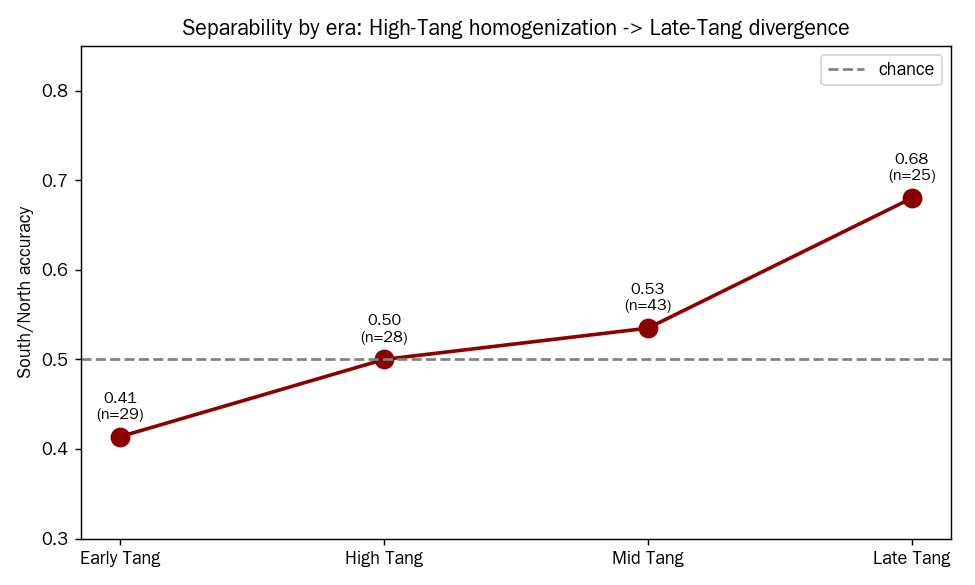}
  \end{minipage}
  \caption{Left: corpus-length control (stable $\Rightarrow$ not a length
  artifact). Right: South/North separability by era.}
  \label{fig:length}
  \label{fig:era}
\end{figure}

We rule out two confounds. \emph{Corpus length} is not responsible: South and
North have nearly identical median corpus sizes, and truncating every poet to a
common budget barely changes accuracy ($0.60 \to 0.59$; Figure~\ref{fig:length}).
\emph{Era} is genuinely associated with region ($\chi^2=11.9$, $p=0.008$:
the Early Tang skews north, the Late Tang skews south, reflecting the southward
migration of literary culture)---but the regional signal persists within eras,
and its strength itself is a finding: South/North is \emph{at chance in the High
Tang} ($0.50$) and \emph{strongest in the Late Tang} ($0.68$;
Figure~\ref{fig:era}). This is quantitative evidence for court-driven
homogenization at the empire's height giving way to regional divergence as
central authority weakened.

\subsection{Reading the model's mistakes as literary history}
\label{sec:errors}
\begin{figure}[t]
  \centering
  \includegraphics[width=0.62\linewidth]{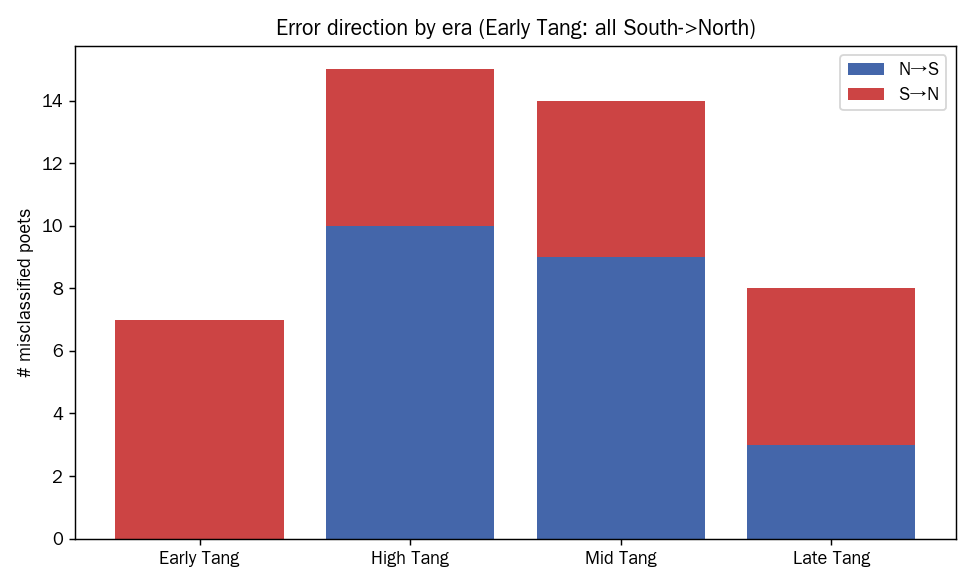}
  \caption{Misclassification direction by era. Early Tang: all South$\to$North.}
  \label{fig:miscls}
\end{figure}

The classifier's confident errors are biographically and historically
meaningful. Broken down by era (Figure~\ref{fig:miscls}), \emph{every} Early-Tang
misclassification is a southern poet read as northern (7 of 7), with no errors in
the opposite direction. Early-Tang southern figures (e.g.\ Yu Shinan \zh{虞世南}, Chu Liang
\zh{褚亮}, He Zhizhang \zh{賀知章}) wrote in the dominant northern court idiom, so the model reads their
verse as northern. The ``error'' thus encodes the homogenizing pull of the
court---corroborating Section~\ref{sec:era}---and illustrates how a model can
generate hypotheses for close reading and literary history.

\subsection{A fair comparison for the transformer}
\begin{figure}[t]
  \centering
  \includegraphics[width=0.7\linewidth]{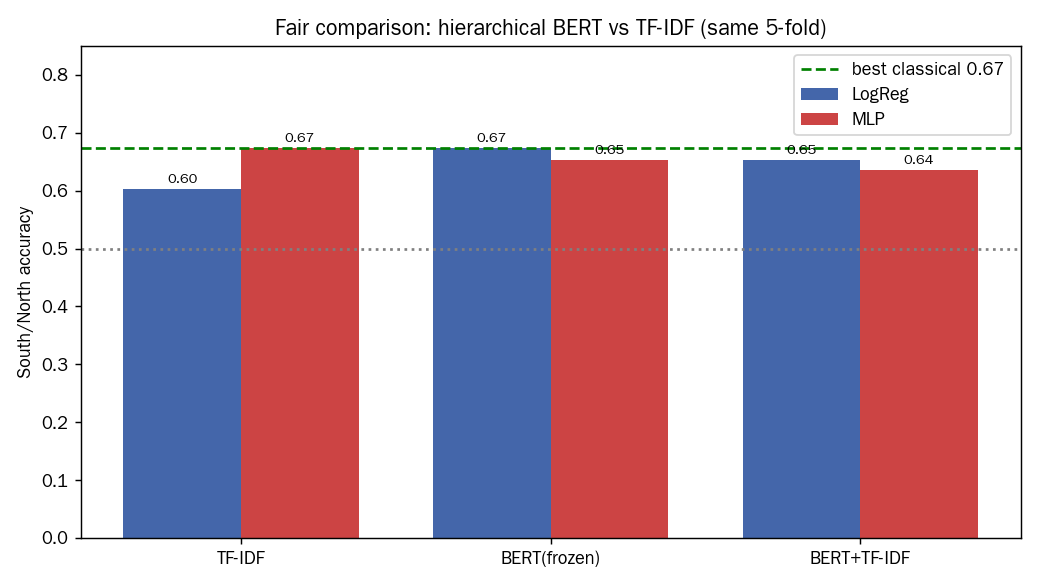}
  \caption{Fair comparison under identical 5-fold CV: hierarchical frozen BERT
  ties TF-IDF; the hybrid does not improve over either.}
  \label{fig:hier}
\end{figure}

Naively fine-tuning GuwenBERT on 250-character fragments and averaging fragment
probabilities yields only $0.62 \pm 0.06$ under grouped cross-validation---but
this handicaps the transformer, which never sees the whole corpus that TF-IDF
reads. We remove the asymmetry with a hierarchical representation: encode each
fragment with a \emph{frozen} GuwenBERT, masked-mean-pool its token states, then
mean-pool all of a poet's fragments into a single vector, evaluated under the
\emph{same} stratified 5-fold protocol. Given the whole corpus, the transformer
now \emph{matches} the best classical model ($0.674$, Figure~\ref{fig:hier});
crucially, combining BERT with TF-IDF adds nothing, indicating that character
$n$-grams already capture the available regional signal. On a corpus of this size
and in a script where the character is the natural unit, the pretrained encoder
offers no information beyond simple lexical statistics.

\section{Discussion}
Methodologically, the study shows that geographic-linguistic analysis---
traditionally confined to spoken dialects---can be applied to a historical
literary corpus, and that interpretable models yield humanistically meaningful
insight rather than black-box prediction. Substantively, our evidence speaks to
the long-running debate over regional schools in Tang poetry: detectable
regional variation persists, concentrated in imagery and lexical choice, even
within an empire that shared a classical education and a court-centered prestige
idiom. The temporal pattern reframes the debate---regional distinctiveness is
not constant but waxes and wanes with the political integration of the dynasty.

\section{Limitations}
The usable sample (357 poets; 242 for South/North) is small, attribution and
geographic labels inherit noise from the source and from CBDB, and the
within-era analyses rest on tens of poets per era. Tonal (\emph{pingze}, \zh{平仄})
features, which require a Middle-Chinese (\zh{中古漢語}) rime dictionary, are approximated only.
The distance-decay effect is suggestive ($p\approx0.09$) and would benefit from
finer (prefecture-level) geocoding and more poets per region.

\section{Conclusion}
A Tang poet's geographic origin leaves a computationally detectable trace in
their verse, carried chiefly by imagery and lexical choice, decaying with
geographic distance and modulated by historical era. Interpretable models, and
even their mistakes, generate concrete hypotheses for literary history---
productive human--machine collaboration in humanistic inquiry. Future work will
extend the temporal analysis and build stylistic similarity networks to ask
whether textual proximity tracks geography or transcends it through literary
influence.

\section*{Data and Code Availability}
The \emph{Complete Tang Poems} text is in the public domain; biographical and
geographic attributions are derived from the China Biographical Database
\citep{cbdb}. All data-processing, modelling, analysis, and figure-generation
code is openly available at
\texttt{https://github.com/ChiShengChen/ctext.org\_-crawler} (subject:
\texttt{gender\_poem\_predictor/}). 
The derived poet-level dataset can be reconstructed from the released scripts.

\section*{Acknowledgements}
We thank the maintainers of the China Biographical Database (CBDB) and the
Chinese Text Project (\zh{中國哲學書電子化計劃}) \citep{ctext} for the resources that made this study
possible, and the developers of GuwenBERT \citep{guwenbert} and
scikit-learn \citep{scikit-learn} for their open-source tools.

\bibliographystyle{plainnat}
\bibliography{references}

@article{scikit-learn,
  title   = {Scikit-learn: Machine Learning in {P}ython},
  author  = {Pedregosa, F. and Varoquaux, G. and Gramfort, A. and Michel, V.
             and Thirion, B. and Grisel, O. and Blondel, M. and Prettenhofer, P.
             and Weiss, R. and Dubourg, V. and Vanderplas, J. and Passos, A.
             and Cournapeau, D. and Brucher, M. and Perrot, M. and Duchesnay, E.},
  journal = {Journal of Machine Learning Research},
  volume  = {12},
  pages   = {2825--2830},
  year    = {2011}
}

@misc{guwenbert,
  title        = {{GuwenBERT}: A Pre-trained Language Model for Classical Chinese},
  author       = {Yan, Ethan},
  year         = {2020},
  howpublished = {\url{https://github.com/Ethan-yt/guwenbert}}
}

@article{liu2019roberta,
  title   = {{RoBERTa}: A Robustly Optimized {BERT} Pretraining Approach},
  author  = {Liu, Yinhan and Ott, Myle and Goyal, Naman and Du, Jingfei and
             Joshi, Mandar and Chen, Danqi and Levy, Omer and Lewis, Mike and
             Zettlemoyer, Luke and Stoyanov, Veselin},
  journal = {arXiv preprint arXiv:1907.11692},
  year    = {2019}
}

@misc{cbdb,
  title        = {China Biographical Database (CBDB)},
  author       = {{Harvard University and Academia Sinica and Peking University}},
  year         = {2024},
  howpublished = {\url{https://projects.iq.harvard.edu/cbdb}}
}

@article{mantel1967,
  title   = {The detection of disease clustering and a generalized regression approach},
  author  = {Mantel, Nathan},
  journal = {Cancer Research},
  volume  = {27},
  number  = {2},
  pages   = {209--220},
  year    = {1967}
}

@article{ctext,
  title   = {The {C}hinese {T}ext {P}roject: A dynamic digital library of premodern Chinese},
  author  = {Sturgeon, Donald},
  journal = {Digital Scholarship in the Humanities},
  volume  = {36},
  pages   = {i101--i112},
  year    = {2021}
}

\end{document}